# Modeling Temporal Dependencies in High-Dimensional Sequences: Application to Polyphonic Music Generation and Transcription

**Nicolas Boulanger-Lewandowski**  BOULANNI@IRO.UMONTREAL.CA
**Yoshua Bengio**  BENGIOY@IRO.UMONTREAL.CA
**Pascal Vincent**  VINCENTP@IRO.UMONTREAL.CA
Dept. IRO, Université de Montréal. Montréal (QC), H3C 3J7, Canada

## Abstract

We investigate the problem of modeling symbolic sequences of polyphonic music in a completely general piano-roll representation. We introduce a probabilistic model based on distribution estimators conditioned on a recurrent neural network that is able to discover temporal dependencies in high-dimensional sequences. Our approach outperforms many traditional models of polyphonic music on a variety of realistic datasets. We show how our musical language model can serve as a symbolic prior to improve the accuracy of polyphonic transcription.

## 1 Introduction

Modeling sequences is an important area of machine learning since many naturally occurring phenomena such as music, speech, or human motion are inherently sequential. Complex sequences are *non-local* in that the impact of a factor localized in time can be delayed by an arbitrarily long time-lag. For example, musical patterns or themes appearing at the beginning of a piece are often repeated towards the end. Recurrent neural networks (RNN) (Rumelhart et al., 1986) incorporate an internal memory that can, in principle, summarize the entire sequence history. This property makes them well suited to represent long-term dependencies, but it is nevertheless a challenge to train them efficiently by gradient-based optimization (Bengio et al., 1994). It was recently shown that training RNNs via Hessian-free (HF) optimization could help reduce these difficulties (Martens & Sutskever, 2011).

Many sequences of interest are over high-dimensional objects, such as images in video, short-term spectra in audio music, tuples of notes in musical scores, or words in text. In these cases, simply predicting the expected value at the next time step given the observed values of the previous time steps is not satisfying. With such high-dimensional objects at each time step, the conditional distribution is very often multi-modal, and we would strongly prefer our models of such sequences to predict the *conditional distribution* of the next time step given previous time steps. For the case of polyphonic music, it is obvious that the occurrence of a particular note at a particular time modifies considerably the probability with which other notes may occur at the same time. In other words, notes appear together in correlated patterns, or *simultaneities*, that cannot be conveniently described by a typical RNN architecture designed for the multiclass classification task, for example, because enumerating all configurations of the variable to predict would be very expensive. This difficulty motivates energy-based models which allow us to express the negative log-likelihood of a given configuration by an arbitrary energy function, among which the restricted Boltzmann machine (RBM) (Smolensky, 1986) has become notorious.

In this context, we wish to exploit the ability of RBMs to represent a complicated distribution for each time step, with parameters that depend on the previous ones, an idea first put forward with the so-called temporal RBM (Taylor et al., 2007; Sutskever & Hinton, 2007) which is trained via a heuristic procedure. Combining the desirable characteristics of RNNs and RBMs has proven to be non-trivial. The recurrent temporal RBM (RTRBM) (Sutskever et al., 2008) is a similar model that allows for exact inference and efficient training by contrastive divergence (CD). Despite its simplicity, this model successfully accounts for several interesting sequences. A similar architecture based on the echo state network was also recently developed (Schrauwen & Buesing, 2009). In this work, we demonstrate that the RTRBM outperforms many traditional models of polyphonic music, and we introduce a generalization of the RTRBM, called the RNN-RBM, that allows more freedom to describe the temporal dependencies involved.

More precisely, we will consider sequences of *symbolic* music, i.e. represented by the explicit timing, pitch, velocity and instrumental information typically contained in a score or a MIDI file rather than more complex, acoustically rich audio signals. Musical models mostly focus on the basic components of western music, harmony and rhythm, and are trained to predict the pattern of notes (simultaneities) to be played together in the next time interval, given the previous ones. Two elements characterize the qualitative performance of a model: temporal dependencies and chord conditional distributions. While most existing models output only monophonic notes along with predefined chords or other reduced-dimensionality representation (e.g. Mozer, 1994; Eck & Schmidhuber, 2002; Paiement et al., 2009), we aim to model unconstrained polyphonic music in the piano-roll representation, i.e. as a binary matrix specifying precisely which notes occur at each time step. Despite ignoring dynamics and other score annotations, this task represents a well-defined framework to improve machine learning algorithms and is directly applicable to polyphonic transcription.

The objective of polyphonic transcription is to determine the underlying notes of a polyphonic audio signal without access to its score. Human experts approach this difficult problem by giving importance to what they expect to hear rather than exclusively to what is present in the actual signal. Most existing transcription algorithms are frame-based and rely exclusively on the audio signal, even though some approaches employ rudimentary musicological constraints (e.g. Li & Wang, 2007). It has long been known that, in the same way that natural language models tremendously improve the performance of speech recognition systems, *musical language models* can improve purely auditive approaches to music information retrieval (Cemgil, 2004). However, combining these two sources of information is not trivial, with the result that temporal smoothing with an HMM is often the only post-processing involved in state-of-the-art transcription (Nam et al., 2011). We will show how to enrich an arbitrary transcription algorithm (under basic assumptions) to include the advice of an expert trained on symbolic sequences. Using our hybrid approach, we can improve transcription accuracy (Bay et al., 2009) much more than the popular HMM approach.

The remainder of the paper is organized as follows. In Sections 2, 3 and 4 we introduce the RBM, the RTRBM and the RNN-RBM architectures. In Section 5 we validate our model on benchmark datasets. In Section 6 we present our results on musical sequences, and we detail our hybrid transcription approach in Section 7.

## 2 Restricted Boltzmann machines

An RBM is an energy-based model where the joint probability of a given configuration of the visible vector $v$ (inputs) and the hidden vector $h$ is:

$$P(v,h) = \exp(-b_v^T v - b_h^T h - h^T W v)/Z \qquad (1)$$

where $b_v$, $b_h$ and $W$ are the model parameters and $Z$ is the usually intractable partition function. When the vector $v$ is given, the hidden units $h_i$ are conditionally independent of one another, and vice versa:

$$P(h_i = 1|v) = \sigma(b_h + Wv)_i \qquad (2)$$
$$P(v_j = 1|h) = \sigma(b_v + W^T h)_j \qquad (3)$$

where $\sigma(x) \equiv (1 + e^{-x})^{-1}$ is the element-wise logistic sigmoid function. The marginalized probability of $v$ is related to the free-energy $F(v)$ by $P(v) \equiv e^{-F(v)}/Z$:

$$F(v) = -b_v^T v - \sum_i \log(1 + e^{b_h + Wv})_i \qquad (4)$$

Inference in RBMs consists of sampling the $h_i$ given $v$ (or the $v_j$ given $h$) according to their conditional Bernoulli distribution (eq. 2). Sampling $v$ from the RBM can be performed efficiently by block Gibbs sampling, i.e. by performing $k$ alternating steps of sampling $h|v$ and $v|h$. The gradient of the negative log-likelihood of an input vector $v^{(l)}$ involves two opposing terms, called the positive and negative phase:

$$\frac{\partial(-\log P(v^{(l)}))}{\partial \Theta} = \frac{\partial F(v^{(l)})}{\partial \Theta} - \frac{\partial(-\log Z)}{\partial \Theta} \qquad (5)$$

where $\Theta \equiv \{b_v, b_h, W\}$. The second term can be estimated by a single sample $v^{(l)*}$ obtained from a $k$-step Gibbs chain starting at $v^{(l)}$:

$$\frac{\partial(-\log P(v^{(l)}))}{\partial \Theta} \simeq \frac{\partial F(v^{(l)})}{\partial \Theta} - \frac{\partial F(v^{(l)*})}{\partial \Theta}. \qquad (6)$$

resulting in the well-known contrastive divergence ($CD_k$) algorithm (Hinton, 2002).

The neural autoregressive distribution estimator (NADE) (Larochelle & Murray, 2011) is a tractable model inspired by the RBM and specializing (with tying constraints) an earlier model for the joint distribution of high-dimensional variables (Bengio & Bengio, 2000). NADE is similar to a fully visible sigmoid belief network in that the conditional probability distribution of a visible unit $v_j$ is expressed as a nonlinear function of $v_k, \forall k < j$. In the following discussion, one can substitute RBMs with NADEs by replacing equation (6) with the exact gradient defined in (Larochelle & Murray, 2011) where the biases are set to $b = v_b^{(t)}$, $c = v_h^{(t)}$. The advantages of a tractable distribution estimator will become obvious when used as part of sequential models.

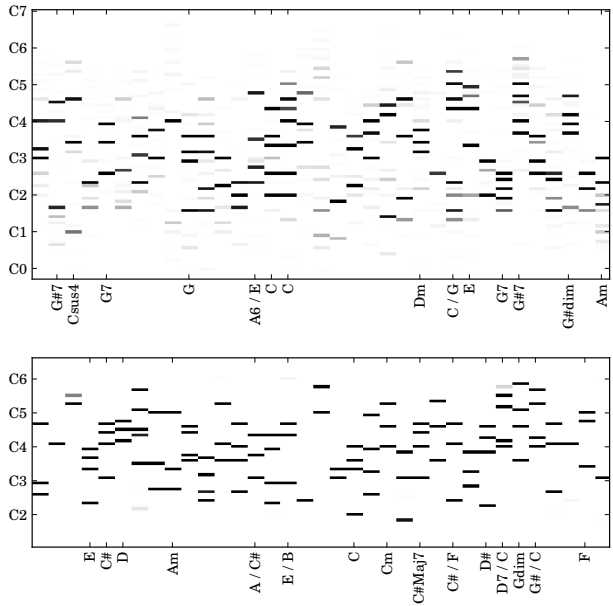

*Figure 1.* Mean-field samples of an RBM trained on the Piano-midi (top) and JSB chorales (bottom) datasets. Each column is a sample vector of notes, with a chord label where the analysis is unambiguous.

Figure 1 presents mean-field samples $P(v_j = 1|h^*)$, where $h^* \sim P(h)$, drawn from RBMs trained on a diverse collection of classical piano music (top) and on the four-part chorales by J. S. Bach (bottom), along with chord labels where the analysis is unambiguous. It is obvious that for the diverse collection, each sample has some room for additional melody notes with probabilities depending on the harmonic context (grey), whereas for JSB chorales, the simultaneities are taken from a more restricted pool and the samples are more clear-cut. This mechanism makes sense musically and the fact that RBMs can adapt to various styles will be useful for the following.

## 3 The RTRBM

The RTRBM (Sutskever et al., 2008) is a sequence of conditional RBMs (one at each time step) whose parameters $b_v^{(t)}, b_h^{(t)}, W^{(t)}$ are time-dependent and depend on the sequence history at time $t$, denoted $\mathcal{A}^{(t)} \equiv \{v^{(\tau)}, \hat{h}^{(\tau)} | \tau < t\}$ where $\hat{h}^{(t)}$ is the mean-field value of $h^{(t)}$. Its graphical structure is depicted in Figure 2(a). The RTRBM is formally defined by its joint probability distribution:

$$P(\{v^{(t)}, h^{(t)}\}) = \prod_{t=1}^{T} P(v^{(t)}, h^{(t)}|\mathcal{A}^{(t)}) \qquad (7)$$

where $P(v^{(t)}, h^{(t)}|\mathcal{A}^{(t)})$ is the joint probability (eq. 1) of the $t^{\text{th}}$ RBM whose parameters are defined below (eq. 8 and 9).

While all the parameters of the RBMs can depend on the previous time steps, we will consider the case where only the biases depend on $\hat{h}^{(t-1)}$:

$$b_h^{(t)} = b_h + W'\hat{h}^{(t-1)} \qquad (8)$$

$$b_v^{(t)} = b_v + W''\hat{h}^{(t-1)} \qquad (9)$$

which gives the RTRBM six parameters: $W, b_v, b_h, W', W'', \hat{h}^{(0)}$. The general case is derived in a similar manner.

While the hidden units $h^{(t)}$ are binary during inference and sampling, it is the *mean-field* value $\hat{h}^{(t)}$ that is transmitted to its successors (see eq. 10). This important distinction makes exact inference of the $\hat{h}^{(t)}$ very easy and improves the efficiency of training (Sutskever et al., 2008):

$$\hat{h}^{(t)} = \sigma(Wv^{(t)} + b_h^{(t)}) = \sigma(Wv^{(t)} + W'\hat{h}^{(t-1)} + b_h) \qquad (10)$$

is obtained directly from equations (2) and (8). Note that equation (10) is exactly the defining equation of a single-layer RNN with hidden units $\hat{h}^{(t)}$.

## 4 The RNN-RBM

The RTRBM can be understood as a sequence of conditional RBMs whose parameters are the output of a deterministic RNN, with the constraint that the hidden units must describe the conditional distributions *and* convey temporal information. This constraint can be lifted by combining a full RNN with distinct hidden units $\hat{h}^{(t)}$ with the RTRBM graphical model as shown in Figure 2(b). We call this model the RNN-RBM. The joint probability distribution of the RNN-RBM is also given by equation (7), but with $\hat{h}^{(t)}$ defined arbitrarily, here as per equation (11).

For simplicity, we consider the RBM parameters to be $W, b_v^{(t)}, b_h^{(t)}$ (i.e. only the biases are variable) and a single-layer RNN (bottom portion of Fig. 2(b)) whose hidden units $\hat{h}^{(t)}$ are only connected to their direct predecessor $\hat{h}^{(t-1)}$ and to $v^{(t)}$ by the relation:

$$\hat{h}^{(t)} = \sigma(W_2 v^{(t)} + W_3 \hat{h}^{(t-1)} + b_{\hat{h}}). \quad (11)$$

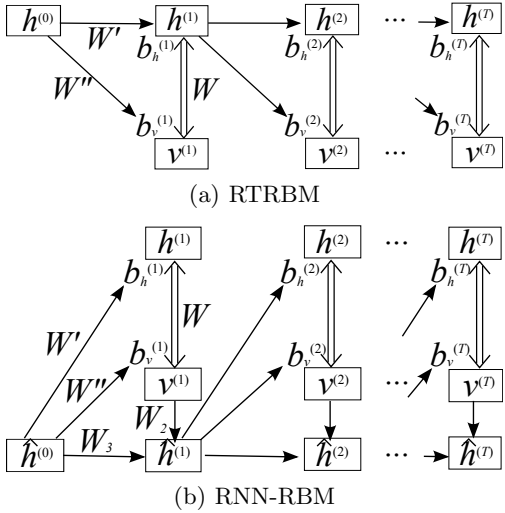

Figure 2. Comparison of the graphical structures of (a) the RTRBM and (b) the single-layer RNN-RBM. Single arrows represent a deterministic function, double arrows represent the stochastic hidden-visible connections of an RBM. The upper half of the RNN-RBM is the RBM stage while the lower half is a RNN with hidden units $\hat{h}^{(t)}$. The RBM biases $b_h^{(t)}$, $b_v^{(t)}$ are a linear function of $\hat{h}^{(t-1)}$.

The RBM portion of the RNN-RBM (upper portion of Fig. 2(b)) is otherwise exactly the same as its RTRBM counterpart. This gives the single-layer RNN-RBM nine parameters: $W, b_v, b_h, W', W'', \hat{h}^{(0)}, W_2, W_3, b_{\hat{h}}$.

The training algorithm is slightly different than for the RTRBM since the mean-field values of the $h^{(t)}$ are now distinct from $\hat{h}^{(t)}$. An iteration of training is based on the following general scheme:

1. Propagate the current values of the hidden units $\hat{h}^{(t)}$ in the RNN portion of the graph using (11),
2. Calculate the RBM parameters that depend on the $\hat{h}^{(t)}$ (eq. 8 and 9) and generate the negative particles $v^{(t)*}$ using $k$-step block Gibbs sampling,
3. Use $CD_k$ to estimate the log-likelihood gradient (eq. 6) with respect to $W$, $b_v^{(t)}$ and $b_h^{(t)}$,
4. Propagate the estimated gradient with respect to $b_v^{(t)}, b_h^{(t)}$ backward through time (BPTT) (Rumelhart et al., 1986) to obtain the estimated gradient with respect to the RNN parameters.

This procedure can be adapted to any RNN architecture and conditional distribution estimator assuming the RNN provides the estimator's parameters (step 2) and can be trained based on a stochastic gradient signal on those parameters (obtained in step 3). The RNN-NADE, obtained by substituting NADEs for RBMs, allows for exact gradient computation.

Note that the single-layer RNN-RBM is a generalization of the RTRBM and reduces to this simpler model by setting $W_2 = W$, $W_3 = W'$ and $b_{\hat{h}} = b_h$ in equations (10) and (11). The RTRBM was not gaining computationally from sharing these connections, hence untying them does not make it slower. In practice, the ability to distinguish between the number of hidden units $h$ and $\hat{h}$ allows to scale RBMs to several hundred hidden units while keeping the RNNs to their (typically smaller) optimal size, improving performance.

### 4.1 Initialization strategies

Initialization strategies based on unsupervised pretraining of each layer have been shown to be important both for supervised and unsupervised training of deep architectures (Bengio, 2009). A recurrent network corresponds to a very deep architecture when unfolded in time, and indeed we find that pretraining can clearly affect the overall performance of both the RTRBM and the RNN-RBM. To ensure the quality of the learned weight matrices, we found that initializing the $W$, $b_v$ and $b_h$ parameters from a trained RBM yields less noisy filters. The hidden-to-bias weights $W', W''$ can then be initialized to small random values, such that the sequential model will initially behave like independent RBMs, eventually departing from that state.

In order to capture better temporal dependencies, we initialize the $W_2, W_3, b_{\hat{h}}, W'', b_v, \hat{h}^{(0)}$ parameters of the RNN-RBM from an RNN trained with the cross-entropy cost:

$$L(\{v^{(t)}\}) = \frac{1}{T} \sum_{t=1}^{T} \sum_{j=1}^{n_v} -v_j^{(t)} \log y_j^{(t)} - (1-v_j^{(t)}) \log(1-y_j^{(t)}) \quad (12)$$

where $y^{(t)} = \sigma(b_v^{(t)})$ and equations (9) and (11) hold. This deterministic objective allows the use of a second-order optimization method for pretraining of the RNN. Note that the RTRBM could use this strategy to initialize $W, W', b_v, b_h, W'', \hat{h}^{(0)}$, but in practice we have found the initialization from an RBM more important.

### 4.2 Details of the BPTT algorithm

Suppose we want to minimize the negative log-likelihood cost $C \equiv -\log P(\{v^{(t)}\})$. The gradient of $C$ with respect to the parameters of the conditional RBMs can be estimated by CD using equations (4) and (6):

$$\frac{\partial C}{\partial b_v^{(t)}} \simeq v^{(t)*} - v^{(t)} \quad (13)$$

$$\frac{\partial C}{\partial W} \simeq \sum_{t=1}^{T} \sigma(Wv^{(t)*} - b_h^{(t)})v^{(t)*\mathrm{T}} - \sigma(Wv^{(t)} - b_h^{(t)})v^{(t)\mathrm{T}} \tag{14}$$

$$\frac{\partial C}{\partial b_h^{(t)}} \simeq \sigma(Wv^{(t)*} - b_h^{(t)}) - \sigma(Wv^{(t)} - b_h^{(t)}). \tag{15}$$

The gradient then back-propagates through the hidden-to-bias parameters (eq. 8 and 9):

$$\frac{\partial C}{\partial W'} = \sum_{t=1}^{T} \frac{\partial C}{\partial b_h^{(t)}} \hat{h}^{(t-1)\mathrm{T}} \tag{16}$$

$$\frac{\partial C}{\partial W''} = \sum_{t=1}^{T} \frac{\partial C}{\partial b_v^{(t)}} \hat{h}^{(t-1)\mathrm{T}} \tag{17}$$

$$\frac{\partial C}{\partial b_h} = \sum_{t=1}^{T} \frac{\partial C}{\partial b_h^{(t)}} \text{ and } \frac{\partial C}{\partial b_v} = \sum_{t=1}^{T} \frac{\partial C}{\partial b_v^{(t)}}. \tag{18}$$

For the single-layer RNN-RBM, the BPTT recurrence relation follows from (11):

$$\frac{\partial C}{\partial \hat{h}^{(t)}} = W_3 \frac{\partial C}{\partial \hat{h}^{(t+1)}} \hat{h}^{(t+1)}(1 - \hat{h}^{(t+1)}) \\ + W' \frac{\partial C}{\partial b_h^{(t+1)}} + W'' \frac{\partial C}{\partial b_v^{(t+1)}} \tag{19}$$

for $0 \leq t < T$ ($\hat{h}^{(0)}$ being a parameter of the model) and $\partial C/\partial \hat{h}^{(T)} = 0$. Formulas for the remaining RNN-RBM parameters are:

$$\frac{\partial C}{\partial b_{\hat{h}}} = \sum_{t=1}^{T} \frac{\partial C}{\partial \hat{h}^{(t)}} \hat{h}^{(t)}(1 - \hat{h}^{(t)}) \tag{20}$$

$$\frac{\partial C}{\partial W_3} = \sum_{t=1}^{T} \frac{\partial C}{\partial \hat{h}^{(t)}} \hat{h}^{(t)}(1 - \hat{h}^{(t)}) \hat{h}^{(t-1)\mathrm{T}} \tag{21}$$

$$\frac{\partial C}{\partial W_2} = \sum_{t=1}^{T} \frac{\partial C}{\partial \hat{h}^{(t)}} \hat{h}^{(t)}(1 - \hat{h}^{(t)}) v^{(t)\mathrm{T}}. \tag{22}$$

## 5 Baseline experiments

In this section, we compare the performance of the RTRBM with the RNN-RBM on two baseline datasets: bouncing balls videos and motion capture data (Sutskever et al., 2008). We use the mean frame-level squared prediction error as a basis of comparison. The prediction of the $t^{\mathrm{th}}$ conditional RBM is performed by 50 steps of block Gibbs sampling starting at $v^{(t-1)}$ and hoping to reconstruct $v^{(t)}$ optimally.

The bouncing ball videos dataset[1] is based on a simulation of balls bouncing in a box (Sutskever & Hinton,

[1] www.cs.utoronto.ca/~ilya/code/2008/RTRBM.tar

2007). The generated videos are of length $T = 128$ and of resolution $15 \times 15$ pixels in the $[0, 1]$ interval, which makes binary RBMs (eq. 1) well suited for this task. With up to 300 hidden units and an initial learning rate of 0.01, we obtain a squared prediction error of 2.11 for the RTRBM and 0.96 for the RNN-RBM, i.e. **less than half the error**. The receptive fields (weights) of the first 48 hidden units $h^{(t)}$ (RNN-RBM) are plotted in Figure 3. Localized edge detectors are apparent in nearly all the learned filters.

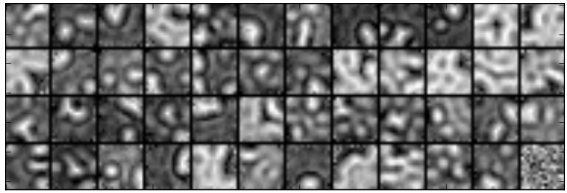

*Figure 3.* Receptive fields of 48 hidden units of an RNN-RBM trained on the bouncing balls dataset. Each square shows the input weights of a hidden unit as an image.

The human motion capture dataset[2] is represented by a sequence of joint angles, translations and rotations of the base of the spine in an exponential-map parameterization (Hsu et al., 2005; Taylor et al., 2007). Since the data consists of 49 real values per time step, we use the Gaussian RBM variant (Welling et al., 2005) for this task. We use up to 450 hidden units and an initial learning rate of 0.001. The mean squared prediction test error is 20.1 for the RTRBM and **reduced substantially to 16.2 for the RNN-RBM**.

## 6 Modeling sequences of polyphonic music

In this section, we show results with main application of interest for this paper: probabilistic modeling of sequences of polyphonic music. We report our experiments on four datasets of varying complexity converted to our input format.

**Piano-midi.de** is a classical piano MIDI archive that was split according to Poliner & Ellis (2007).
**Nottingham** is a collection of 1200 folk tunes[3] with chords instantiated from the ABC format.
**MuseData** is an electronic library of orchestral and piano classical music from CCARH[4].
**JSB chorales** refers to the entire corpus of 382 four-part harmonized chorales by J. S. Bach with the split of Allan & Williams (2005).

[2] people.csail.mit.edu/ehsu/work/sig05stf
[3] ifdo.ca/~seymour/nottingham/nottingham.html
[4] www.musedata.org

Each dataset contains at least 7 hours of polyphonic music and the total duration is approximately 67 hours. The polyphony (number of simultaneous notes) varies from 0 to 15 and the average polyphony is 3.9. We use an input of 88 binary visible units that span the whole range of piano from A0 to C8 and temporally aligned on an integer fraction of the beat (quarter note). Consequently, pieces with different time signatures will not have their measures start at the same interval. Although it is not strictly necessary, learning is facilitated if the sequences are transposed in a common tonality (e.g. C major/minor) as preprocessing.

In addition to the models previously described, we evaluate the following commonly used methods:

- The simplest baseline model consists in outputting a Gaussian density centered on the previous frame $\mu = v^{(t-1)}$ and learned covariance $\Sigma$.
- N-grams simulate the evolution of note simultaneities as an $(N-1)^{\text{th}}$-order Markov chain. We use add-$p$ or Gaussian smoothing and back-off.
- Note N-grams model each note independently by a binary N-gram, possibly with shared parameters (IID).
- An interesting model for chorales harmonisation (Allan & Williams, 2005) has been adapted to serve as a generative model. It can only be evaluated on the JSB chorales dataset.
- The 'random fields' approach of Lavrenko & Pickens (2003) is a type of fully visible sigmoid belief network with learned connectivity.
- Other common methods include Gaussian mixture models (GMM), hidden Markov models (HMM) using GMM indices as their state, and multilayer perceptrons (MLP) with the last $n$ time steps as input.

The log-likelihood (LL) and expected frame-level accuracy (ACC) (Bay et al., 2009) of the symbolic models are presented in Table 1. We estimate the partition function of each conditional RBM by 100 runs of annealed importance sampling (Salakhutdinov & Murray, 2008). We make a few key observations:

- The complexity of the dataset, such as the simplistic chord accompaniment of Nottingham and the redundant style of four-part chorales by a single composer, in comparison with diverse piano and orchestral music, is clearly reflected in the obtained log-likelihoods and accuracies.
- N-gram models (optimal $N^* = 2$) perform reasonably well for simple datasets but fail in more realistic settings due to the increased data sparsity. In this case, note N-grams ($N^* \in [8, 14]$) are a better alternative albeit ignoring harmonic dependencies. This inherent trade-off in traditional polyphonic music models can be addressed robustly by the RNN-based models, that perform better on a range of datasets.
- The harmonisation model of Allan & Williams (2005), tailored to the specific style of four-part chorales, requires annotated harmonic symbols and yet performs relatively poorly compared to our best performer. Similarly to the GMM + HMM, this model is penalized by the limited history of the HMM and by the difficulty to generalize to new chord voicings in a principled manner.
- In accordance with earlier results (Martens & Sutskever, 2011), the use of HF significantly helps the density estimation and prediction performance of RNNs (eq. 12) which would otherwise perform worse than simpler MLPs. This motivates our strategy of pretraining the RNN layer of an RNN-RBM via HF.
- In addition to the distinct recurrent hidden units $\hat{h}^{(t)}$ that convey temporal information more freely, and the fact that suitable learning rates can be specified differently for the RNN and the RBM parts, pretraining the $W_2$, $W_3$ and $b_{\hat{h}}$ parameters can have the most impact on the RNN-RBM prediction performance. Figure 4 clearly demonstrates the importance of pretraining and finetuning the RNN and the additional advantage of using HF.
- Although frame-level NADEs are slightly less powerful than RBMs, their desirable properties make the combined RNN-NADE model the most robust distribution estimator. We believe this is due to their tractable distribution, for two reasons. First, CD may not be ideally suited for conditional RBMs with slowly-mixing Gibbs chains (Mnih et al., 2011), a non-issue for exact-gradient models. Secondly, the joint sequential model, and not only the RNN portion, can benefit from second-order optimization as can be seen from the last two rows of Table 1.

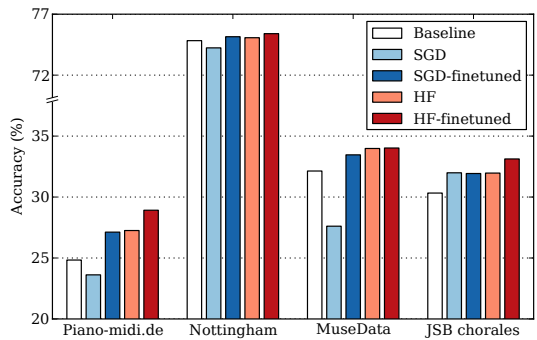

*Figure 4.* Effect of SGD and HF pretraining on the RNN-RBM symbolic prediction performance. All strategies except the baseline involve pretraining.

Modeling Temporal Dependencies in High-Dimensional SequencesTable 1. Log-likelihood and expected accuracy for various musical models in the symbolic prediction task. The double line separates frame-level models (above) and models with a temporal component (below).

| Model | Piano-midi.de LL | ACC % | Nottingham LL | ACC % | MuseData LL | ACC % | JSB chorales LL | ACC % |
|---|---|---|---|---|---|---|---|---|
| Random | -61.00 | 3.35 | -61.00 | 4.53 | -61.00 | 3.74 | -61.00 | 4.42 |
| 1-Gram (Add-$p$) | -27.64 | 4.85 | -5.94 | 22.76 | -19.03 | 6.67 | -12.22 | 16.80 |
| 1-Gram (Gaussian) | -10.79 | 6.04 | -5.30 | 21.31 | -10.15 | 7.87 | -7.56 | 17.41 |
| Note 1-Gram | -11.05 | 5.80 | -10.25 | 19.87 | -11.51 | 7.72 | -11.06 | 15.25 |
| Note 1-Gram (IID) | -12.90 | 2.51 | -16.24 | 3.56 | -14.06 | 2.82 | -15.93 | 3.51 |
| GMM | -15.84 | 5.08 | -7.87 | 22.62 | -12.20 | 7.37 | -11.90 | 15.84 |
| RBM | -10.17 | 5.63 | -5.25 | 5.81 | -9.56 | 8.19 | -7.43 | 4.47 |
| NADE | -10.28 | 5.82 | -5.48 | 22.67 | -10.06 | 7.65 | -7.19 | 17.88 |
| Previous + Gaussian | -12.48 | 25.50 | -8.41 | 55.69 | -12.90 | 25.93 | -19.00 | 18.36 |
| N-Gram (Add-$p$) | -46.04 | 7.42 | -6.50 | 63.45 | -35.22 | 10.47 | -29.98 | 24.20 |
| N-Gram (Gaussian) | -12.22 | 10.01 | -3.16 | 65.97 | -10.59 | 16.15 | -9.74 | 28.79 |
| Note N-Gram | -7.50 | 26.80 | -4.54 | 62.49 | -7.91 | 26.35 | -10.26 | 20.34 |
| GMM + HMM | -15.30 | 7.91 | -6.17 | 59.27 | -11.17 | 13.93 | -11.89 | 19.24 |
| (Allan & Williams, 2005) | – | – | – | – | – | – | -9.24 | 16.32 |
| (Lavrenko & Pickens, 2003) | -9.05 | 18.37 | -5.44 | 55.34 | -9.87 | 18.39 | -8.78 | 22.93 |
| MLP | -8.13 | 20.29 | -4.38 | 63.46 | -7.94 | 25.68 | -8.70 | 30.41 |
| RNN | -8.37 | 19.33 | -4.46 | 62.93 | -8.13 | 23.25 | -8.71 | 28.46 |
| RNN (HF) | -7.66 | 23.34 | -3.89 | 66.64 | -7.19 | 30.49 | -8.58 | 29.41 |
| RTRBM | -7.36 | 22.99 | -2.62 | 75.01 | -6.35 | 30.85 | -6.35 | 30.17 |
| RNN-RBM | **-7.09** | **28.92** | **-2.39** | **75.40** | -6.01 | **34.02** | -6.27 | **33.12** |
| RNN-NADE | -7.48 | 20.69 | -2.91 | 64.95 | -6.74 | 24.91 | -5.83 | 32.11 |
| RNN-NADE (HF) | **-7.05** | 23.42 | **-2.31** | 71.50 | **-5.60** | 32.60 | **-5.56** | 32.50 |

We evaluate our models qualitatively by generating sample sequences, provided on the authors' website[5], and discussed here. While note correlations are obviously neglected in the simpler models (sequence 2), RBM-based models learned basic harmony rules (sequence 3), melody lines (sequences 4, 8) and local temporal coherence (sequence 5). However, long-term structure and musical meter remain elusive.

## 7 Polyphonic transcription

Multiple fundamental frequency ($f_0$) estimation, or polyphonic transcription, consists in estimating the audible note pitches in the signal at 10 ms intervals without tracking note contours. We combine our polyphonic sequence models with the acoustic model of Nam et al. (2011) in order to demonstrate a practical application of the sequence models. Their model was adapted for multiple instruments, and it can be generalized to any method that can score hypothetical combinations of $f_0$ for a given time frame.

At each time frame, the Nam et al. (2011) algorithm outputs independent probabilities that each note is present and reports every note with probability $p \geq 0.5$. To incorporate our symbolic model prediction $P_s(v^{(t)}|\mathcal{A}^{(t)})$, we consider the $k$ most promising $f_0$ candidates ($k = 7$) from the acoustic model $P_a(v^{(t)})$ and jointly evaluate all combinations of $M$ candidates $\forall M \leq k$ by the following cost function:

$$C = -\log P_a(v^{(t)}) - \alpha \log P_s(v^{(t)}|\tilde{\mathcal{A}}^{(t)}) \qquad (23)$$

where $\tilde{\mathcal{A}}^{(t)}$ is the approximate sequence history constructed from the $f_0$ estimated so far in at least half the audio frames corresponding to each past symbolic time step[6]. This corresponds to a product of experts where the hyperparameter $\alpha$ is the confidence coefficient of our symbolic predictor. If our algorithm is run on audio signals without preprocessing, tempo tracking must be performed first. Since the symbolic models describe only fixed tonality pieces, a first audio-only pass is needed to transpose the estimated $f_0$ in the correct tonality. Once the optimal $f_0$ estimates have been determined, HMM smoothing can still filter out spurious results and enhance onset accuracies.

Digital audio has been generated for the four datasets and we report in Figure 5 the frame-level transcription accuracy of the Nam et al. (2011) algorithm, either alone, after HMM smoothing, or using our best performing model as a symbolic prior. We observe an improvement in absolute accuracy between 1.3% and 10% over the HMM approach. It can be seen easily that an HMM with emission probabilities $P_a(v^{(t)})$ is equivalent to equation (23) with a note 2-gram symbolic model, one time step per audio frame and $\alpha = 1$. It is therefore unsurprising that the advantage of our search algorithm decreases when the note N-gram already performs well, e.g. for Piano-midi.de (Table 1). However, the HMM allows for a *global* search of the most likely $f_0$ (the Viterbi path), whereas our algorithm requires a greedy chronological search, a limitation we are currently working to address.

---

[5] www-etud.iro.umontreal.ca/~boulanni/icml2012

[6] This can create a 'snowball' effect where accurate baseline transcriptions form accurate $\tilde{\mathcal{A}}^{(t)}$ estimates, resulting in more relevant symbolic predictions $P_s(v^{(t)}|\tilde{\mathcal{A}}^{(t)})$, which in turn improve the final transcription.

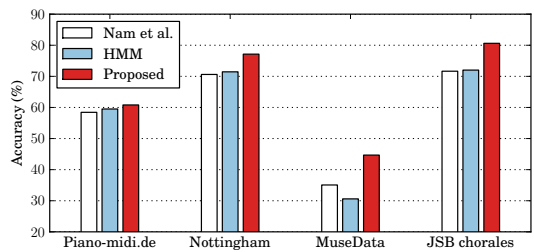

*Figure 5.* Frame-level transcription accuracy of the Nam et al. (2011) model either alone, after HMM smoothing or with our best performing model as a symbolic prior.

## 8  Conclusions

We presented an RNN-based model that can learn harmonic and rhythmic probabilistic rules from polyphonic music scores of varying complexity, substantially better than popular methods in music information retrieval. We showed that different strategies related to the description of temporal dependencies can improve prediction accuracy of such models. While longer-term musical structure remains elusive in our unconstrained representation, our model can immediately serve as a symbolic prior for polyphonic transcription, clearly improving the state of the art in this area.

## Acknowledgments

The authors would like to thank NSERC, CIFAR and the Canada Research Chairs for funding, and Compute Canada/Calcul Québec for computing resources.

## References

Allan, M. and Williams, C.K.I. Harmonising chorales by probabilistic inference. In *NIPS 17*, pp. 25–32, 2005.

Bay, M., Ehmann, A.F., and Downie, J.S. Evaluation of multiple-F0 estimation and tracking systems. In *ISMIR*, 2009.

Bengio, Y. Learning deep architectures for AI. *Foundations and Trends in Machine Learning*, 2(1):1–127, 2009.

Bengio, Y. and Bengio, S. Modeling high-dimensional discrete data with multi-layer neural networks. In *NIPS 12*, pp. 400–406, 2000.

Bengio, Y., Simard, P., and Frasconi, P. Learning long-term dependencies with gradient descent is difficult. *IEEE Trans. on Neural Networks*, 5(2):157–166, 1994.

Cemgil, A.T. *Bayesian music transcription*. PhD thesis, Radboud University Nijmegen, 2004.

Eck, D. and Schmidhuber, J. Finding temporal structure in music: Blues improvisation with LSTM recurrent networks. In *NNSP*, pp. 747–756, 2002.

Hinton, G.E. Training products of experts by minimizing contrastive divergence. *Neural Computation*, 14(8): 1771–1800, 2002.

Hsu, E., Pulli, K., and Popović, J. Style translation for human motion. In *SIGGRAPH*, pp. 1082–1089, 2005.

Larochelle, H. and Murray, I. The neural autoregressive distribution estimator. *JMLR: W&CP*, 15:29–37, 2011.

Lavrenko, V. and Pickens, J. Polyphonic music modeling with random fields. In *ACM MM*, pp. 120–129, 2003.

Li, Y. and Wang, D.L. Pitch detection in polyphonic music using instrument tone models. *ICASSP*, 2:481, 2007.

Martens, J. and Sutskever, I. Learning recurrent neural networks with Hessian-free optimization. In *ICML 28*, 2011.

Mnih, V., Larochelle, H., and Hinton, G. Conditional restricted Boltzmann machines for structured output prediction. In *UAI 27*, 2011.

Mozer, M.C. Neural network music composition by prediction. *Connection Science*, 6(2):247–280, 1994.

Nam, J., Ngiam, J., Lee, H., and Slaney, M. A classification-based polyphonic piano transcription approach using learned feature representations. In *ISMIR*, 2011.

Paiement, J.F., Bengio, S., and Eck, D. Probabilistic models for melodic prediction. *Artificial Intelligence*, 173 (14):1266–1274, 2009.

Poliner, G.E. and Ellis, D.P.W. A discriminative model for polyphonic piano transcription. *JASP*, 2007(1):154–164, 2007.

Rumelhart, D.E., Hinton, G.E., and Williams, R.J. Learning internal representations by error propagation. In *Parallel Dist. Proc.*, pp. 318–362. MIT Press, 1986.

Salakhutdinov, R. and Murray, I. On the quantitative analysis of deep belief networks. In *ICML 25*, 2008.

Schrauwen, B. and Buesing, L. A hierarchy of recurrent networks for speech recognition. In *NIPS 21*, 2009.

Smolensky, P. Information processing in dynamical systems: Foundations of harmony theory. In *Parallel Dist. Proc.*, pp. 194–281. MIT Press, 1986.

Sutskever, I. and Hinton, G.E. Learning multilevel distributed representations for high-dimensional sequences. In *AISTATS*, pp. 544–551, 2007.

Sutskever, I., Hinton, G., and Taylor, G. The recurrent temporal restricted Boltzmann machine. In *NIPS 20*, pp. 1601–1608, 2008.

Taylor, G.W., Hinton, G.E., and Roweis, S.T. Modeling human motion using binary latent variables. In *NIPS 19*, pp. 1345, 2007.

Welling, M., Rosen-Zvi, M., and Hinton, G. Exponential family harmoniums with an application to information retrieval. In *NIPS 17*, pp. 1481–1488, 2005.